\PassOptionsToPackage{usenames,dvipsnames,table,hyperref}{xcolor}
\documentclass[11pt,a4paper]{article}
\usepackage{emnlp-ijcnlp-2019}
\usepackage{times}
\usepackage{latexsym}
\usepackage{url}

\usepackage{tabularx}
\usepackage{times}
\usepackage{latexsym}
\usepackage{amssymb}
\usepackage{times}
\usepackage{latexsym}
\usepackage{amsmath}
\usepackage{multirow}
\usepackage{amsfonts}
\usepackage{booktabs}
\usepackage{rotating}
\usepackage{subfig}
\usepackage{array}
\usepackage{booktabs}
\usepackage{enumitem}
\usepackage{makecell}
\usepackage{xcolor}
\usepackage{caption}
\usepackage{array}
\usepackage[table]{xcolor}
\usepackage[toc,page]{appendix}
\usepackage{multirow}
\usepackage{xspace} 
\usepackage{makecell}

\definecolor{vlgray}{gray}{0.6}
\usepackage{floatrow}
\usepackage{amsmath}
\usepackage{etoolbox}

\newfloatcommand{capbtabbox}{table}[][\FBwidth]

\newcommand{\thickbar}[1]{\mathbf{\bar{\text{$#1$}}}}

\newtoggle{draft}
\toggletrue{draft}
\iftoggle{draft}{
\newcommand{\varun}[1]{\textcolor{Cyan}{[#1 \textsc{--varun}]}}
\newcommand{\dk}[1]{\textcolor{Maroon}{[#1 \textsc{--dk}]}}
\newcommand{\ed}[1]{\textcolor{Maroon}{[#1 \textsc{--ed}]}}
\newcommand{\taehee}[1]{\textcolor{BurntOrange}{[#1 \textsc{--taehee}]}}

}{
\newcommand{\varun}[1]{}
\newcommand{\dk}[1]{}
\newcommand{\ed}[1]{}
\newcommand{\taehee}[1]{}
}

\def\BState{\State\hskip-\ALG@thistlm}

\usepackage{etoolbox}
\preto\align{\par\nobreak\normalsize\noindent}
\expandafter\preto\csname align*\endcsname{\par\nobreak\normalsize\noindent}

\newcommand{\methodpastel}{\textsc{PASTEL}\xspace}
\newcolumntype{P}[1]{>{\centering\arraybackslash}p{#1}}
\newcolumntype{C}[1]{>{\centering}p{#1}}
\newcolumntype{R}{>{\raggedleft\arraybackslash}X}

\aclfinalcopy 

\title{
(Male, Bachelor) and (Female, Ph.D) have different connotations: Parallelly Annotated Stylistic Language Dataset with Multiple Personas}

\author{
Dongyeop Kang \quad Varun Gangal \quad Eduard Hovy\\
Carnegie Mellon University, Pittsburgh, PA, USA \\
{\tt $\{$dongyeok,vgangal,hovy$\}$@cs.cmu.edu   }
}

\date{}

\begin{document}
\maketitle

\begin{abstract}
Stylistic variation in text needs to be studied with different aspects including the writer's personal traits, interpersonal relations, rhetoric, and more.
Despite recent attempts on computational modeling of the variation, the lack of parallel corpora of style language makes it difficult to systematically control the stylistic change as well as evaluate such models.
We release \methodpastel, the parallel and annotated stylistic language dataset, that contains $\approx41$K parallel sentences ($8.3$K parallel stories) annotated across different personas.
Each persona has different styles in conjunction: gender, age, country, political view, education, ethnic, and time-of-writing.
The dataset is collected from human annotators with solid control of input denotation: not only preserving original meaning between text, but promoting stylistic diversity to annotators.
We test the dataset on two interesting applications of style language, where \methodpastel helps design appropriate experiment and evaluation. 
First, in predicting a target style (e.g., male or female in gender) given a text, multiple styles of \methodpastel make other external style variables controlled (or fixed), which is a more accurate experimental design.
Second, a simple supervised model with our parallel text outperforms the unsupervised models using non-parallel text in style transfer.
Our dataset is publicly available\footnote{\url{https://github.com/dykang/PASTEL}}.
\end{abstract}

\section{Introduction}
\citet{hovy1987generating} claims that appropriately varying the style of text often conveys more information than is contained in the literal meaning of the words. 
He defines the roles of styles in text variation by pragmatics aspects (e.g., relationship between them) and rhetorical goals (e.g., formality), and provides example texts of how they are tightly coupled in practice. 
Similarly, \citet{biber1991variation} categorizes components of conversational situation by participants' characteristics such as  their roles, personal characteristics, and group characteristics (e.g., social class). 
Despite the broad definition of style, this work mainly focuses on one specific aspect of style, \textit{pragmatics aspects in group characteristics of speakers}, which is also called \textit{persona}.
Particularly, we look at multiple types of group characteristics in conjunction, such as gender, age, education level, and more.

Stylistic variation in text primarily manifest themselves at the different levels of textual features: lexical features (e.g., word choice), syntactic features (e.g., preference for the passive voice) and even pragmatics, while preserving the original meaning of given text \cite{dimarco1990accounting}. 
Connecting such textual features to someone's persona is an important study to understand stylistic variation of language.
For example, do highly educated people write longer sentences \cite{bloomfield1927literate}? 
Are Hispanic and East Asian people more likely to drop pronouns \cite{white1985pro}? 
Are elder people likely to use lesser anaphora \cite{ulatowska1986disruption}?

To computationally model a meaning-preserved variance of text across styles, many recent works have developed systems that transfer styles \cite{reddy2016obfuscating,hu2017toward,prabhumoye2018style} or profiles authorships from text \cite{verhoeven2014clips,koppel2009computational,stamatatos2018overview} without parallel corpus of stylistic text.
However, the absence of such a parallel dataset makes it difficult both to systematically learn the textual variation of multiple styles as well as properly evaluate the models.


In this paper, we propose a \textit{large scale, human-annotated, parallel stylistic dataset} called \methodpastel, with focus on \textit{multiple types of personas in conjunction}.
Ideally, annotations for a parallel style dataset should preserve the original meaning (i.e., \textit{denotation}) between reference text and stylistically transformed text, while promoting diversity for annotators to allow their own styles of persona (i.e., \textit{connotation}).
However, if annotators are asked to write their own text given a reference sentence, they may simply produce arbitrarily paraphrased output which does not exhibit a stylistic diversity. 
To find such a proper input setting for data collection, we conduct a denotation experiment in \S\ref{sec:denotation}.
\methodpastel is then collected by crowd workers based on the most effective input setting that balances both meaning preservation and diversity  metrics (\S\ref{sec:dataset}).


\methodpastel includes stylistic variation of text at two levels of parallelism: $\approx$8.3K annotated, parallel stories and $\approx$41K annotated, parallel sentences, where each story has five sentences and has $2.63$ annotators on average.
Each sentence or story has the seven types of persona styles in conjunction: gender, age, ethnics, countries to live, education level, political view, and time of the day.

In \S\ref{sec:exp}, we introduce two interesting applications of style language using \methodpastel: controlled style classification and supervised style transfer.
The former application predicts a category (e.g., male or female) of target style (i.e., gender) given a text.
Multiplicity of persona styles in \methodpastel makes other style variables controlled (or fixed) except the target, which is a more accurate experimental design.
In the latter, contrast to the unsupervised style transfer using non-parallel corpus, simple supervised models with our parallel text in \methodpastel achieve better performance, being evaluated with the parallel, annotated text.

We hope \methodpastel sheds light on the study of stylistic language variation in developing a solid model as well as evaluating the system properly.


\section{Related Work}

Transferring styles between text has been studied with and without parallel corpus:


\textit{Style transfer without parallel corpus: }
Prior works transfer style between text on single type of style aspect such as sentiment~\cite{fu2017style,shen2017style,hu2017toward}, gender~\cite{reddy2016obfuscating}, political orientation~\cite{prabhumoye2018style}, and two conflicting corpora (e.g, paper and news~\cite{han2017unsupervised}, or real and synthetic reviews~\cite{lipton2015generative}).
They use different types of generative models in the same way as style transfer in images, where meaning preservation is not controlled systematically.
\citet{prabhumoye2018style} proposes back-translation to get a style-agnostic sentence representation. 
However, they lack parallel ground truth for evaluation and present limited evaluation for meaning preservation.

\textit{Style transfer with parallel corpus: }
Few recent works use parallel text for style transfer between modern and Shakespearean text \cite{jhamtani2017shakespearizing}, sarcastic and literal tweets \cite{peled2017sarcasm}, and formal and informal text \cite{heylighen1999formality,rao2018dear}.
Compared to these, we aim to understand and demonstrate style variation owing to multiple demographic attributes.

Besides the style transfer, other applications using stylistic features have been studied such as poetry generation~\cite{ghazvininejad2017hafez}, stylometry with demographic information~\cite{verhoeven2014clips}, modeling style bias~\cite{vogel2012he} and modeling biographic attributes~\cite{garera2009modeling}. 
A series of works by \cite{wild,koppel2009computational,argamon2009automatically,koppel2014determining} and their shared tasks \cite{stamatatos2018overview} show huge progress on author profiling and attribute classification tasks.
However, none of the prior works have collected a stylistic language dataset to have multiple styles in conjunction, parallely annotated by a human.
The multiple styles in conjunction in \methodpastel enable an appropriate experiment setting for controlled style classification task in Section~\ref{sec:classification}.

\section{Denotation Experiment}\label{sec:denotation}

\begin{figure*}[htbp!]
\vspace{0mm}
\begin{floatrow}
\capbtabbox{%
\small
\begin{tabularx}{1.0\linewidth}{@{}p{2.15cm}@{\hskip 2mm}X@{}}
\toprule
\textbf{Denotation}: & \textbf{Produced sentences}:\\\midrule
\textit{single ref. sentence} & the old door with wood was the only direction to the courtyard \\\midrule
\textit{story(imgs)} & The old wooden door in the stonewall looks like a portal to a fairy tale.  \\\midrule
\textit{story(imgs.+keyw words)} & Equally so, he is intrigued by the heavy wooden door in the courtyard. \\\midrule
\multicolumn{2}{@{}l}{\textbf{Reference sentence}:}\\
\multicolumn{2}{@{}l}{the old wooden door was only one way into the courtyard.}  \\
\bottomrule
\end{tabularx}\vspace{0mm}
}{%
  \caption{\label{tab:example_variation} Textual variation across different denotation settings. Each sentence is produced by a same annotator. Note that providing reference sentence increases fidelity to the reference while decreases diversity.}%
}
\ffigbox{%
\includegraphics[trim={0mm 8.00cm 13.5cm 0mm},clip,width=1.07\linewidth]{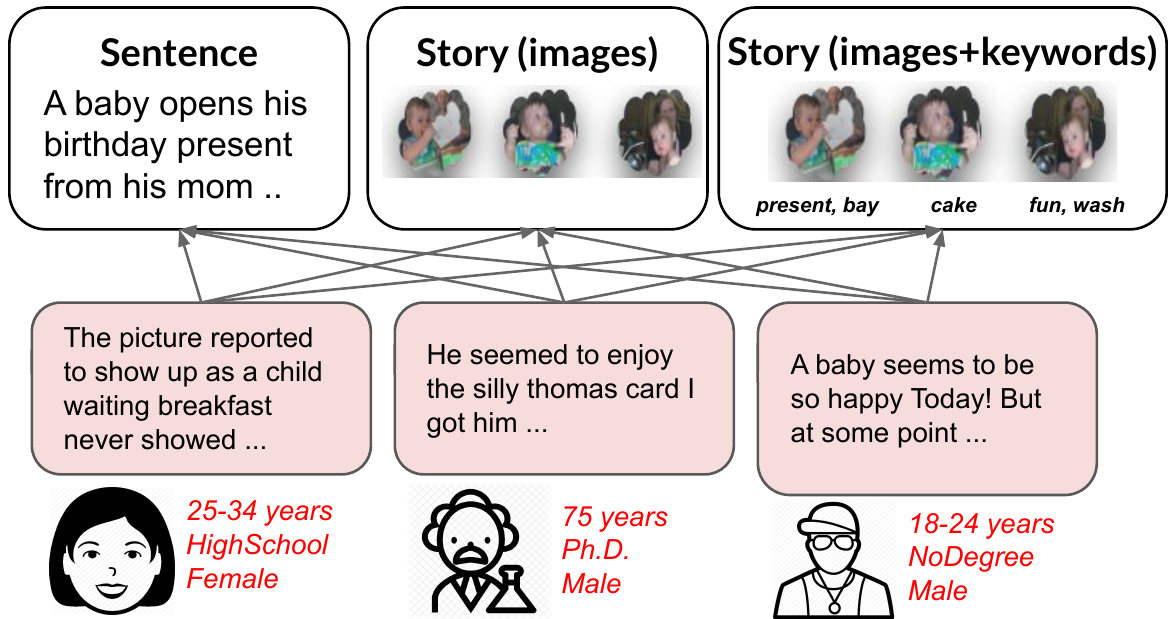}
}{%
  \caption{
  \label{fig:denotation_experiment}
Denotation experiment finds the best input setting for data collection, that preserves meaning but diversifies styles among annotators with different personas. }%
}
\end{floatrow}
\vspace{0mm}
\end{figure*}



We first provide a preliminary study to find the best input setting (or denotation) for data collection to balance between two trade-off metrics: meaning preservation and style diversity.

\subsection{Preliminary Study}
Table~\ref{tab:example_variation} shows output texts produced by annotators given different input denotation settings.
The basic task is to provide an input denotation (e.g., a sentence only, a sequence of images) and then ask them to reproduce text maintaining the meaning of the input but with their own persona. 

For instance, if we provide a \textit{single reference sentence}, annotators mostly repeat the input text with a little changes of the lexical terms. 
This setup mostly preserves the meaning by simply paraphrasing the sentence, but annotators' personal style does not reflect the variation.
With a \textit{single image}, on the other hand, the outputs produced by annotators tend to be diverse. However, the image can be explained with a variety of contents, so the output meaning can drift away from the reference sentence.

If a series of consistent images (i.e., a story) is given, we expect a stylistic diversity can be more narrowed down, by grounding it to a specific event or a story.
In addition to that, some keywords added to each image of a story help deliver more concrete meaning of content as well as the style diversity.

\begin{table*}[ht!]
\small
\vspace{0mm}
\begin{center}
\caption{\label{tab:denotation} Denotation experiment to find the best input setting (i.e., meaning preserved but stylistically diverse). \textbf{story-level} measures the metrics for five sentences as a story, and \textbf{sentence-level} per individual sentence.
Note that \textit{single reference sentence} setting only has sentence level. For every metrics in both meaning preservation and style diversity, the higher the better. The \textbf{bold} number is the highest, and the \underline{underlined} is the second highest. 
\vspace{0mm}
}
\centering
\begin{tabular}{@{}c@{}|l|c|cc@{}}
\toprule
 &\textit{denotation settings}&  \multicolumn{1}{|c|}{\textbf{Style Diversity}} &  \multicolumn{2}{|c}{\textbf{Meaning Preservation}}   \\\midrule
& &	E(GM) & METEOR & VectorExtrema\\
\midrule
\parbox[t]{3mm}{\multirow{5}{*}{\rotatebox[origin=c]{90}{{\footnotesize\textbf{sentence-level}}}}} 
&\textit{single ref. sentence} &  \underline{2.98} &  \textbf{0.37} & \textbf{0.70}\\
&\textit{\textbf{story}(images)}&  2.86 &  0.07 & 0.38\\
&\textit{\textbf{story}(images) + global keywords}&  2.85 &0.07 & 0.39\\
&\textit{\textbf{story}(images + local keywords)}& \textbf{3.07} &   0.17 &\underline{0.53}\\
&\textit{\textbf{story}(images + local keywords + ref. sentence)}&  2.91 & \underline{0.21} & 0.43\\
\midrule
\parbox[t]{3mm}{\multirow{4}{*}{\rotatebox[origin=c]{90}{{\footnotesize\textbf{story-level}}}}} 
&\textit{\textbf{story}(images)}&  4.43  & 0.1 & 0.4\\
&\textit{\textbf{story}(images) + global keywords}&  4.43 &  0.1 &  0.42\\
&\textit{\textbf{story}(images + local keywords)}& \textbf{4.58}  & \underline{0.19} & \textbf{0.55}\\
&\textit{\textbf{story}(images + local keywords + ref. sentence)}& \underline{4.48} &\textbf{0.22} & \underline{0.44}\\
\bottomrule
\end{tabular}
\end{center}
\vspace{0mm}
\end{table*}

\subsection{Experimental Setup}
In order to find the best input setting that preserves meaning as well as promotes a stylistic diversity, we conduct a denotation experiment as described in Figure \ref{fig:denotation_experiment}.
The experiment is a subset of our original dataset, which have only 100 samples of annotations.

A basic idea behind this setup is to provide (1) a perceptually common denotation via sentences or images so people share the same context (i.e., denotation) given, (2) a series of them as a ``story'' to limit them into a specific event context, and (3) two modalities (i.e., text and image) for better disambiguation of the context by grounding them to each other.

We test five different input settings\footnote{Other settings like \textit{Single reference image} are tested as well, but they didn't preserve the meaning well.}: 
\textit{Single reference sentence}, \textit{\textbf{Story} (images)}, \textit{\textbf{Story} (images) + global keywords}, \textit{\textbf{Story} (images + local keywords)}, and \textit{\textbf{Story} (images + local keywords + ref. sentence)}.

For the keyword selection, we use RAKE algorithm~\cite{rose2010automatic} to extract keywords and rank them for each sentence by the output score.
Top five uni/bigram keywords are chosen at each story, which are called \textit{global keywords}.
On the other hand, another top three uni/bigram keywords are chosen at each image/sentence in a story, which are called \textit{local keywords}.
Local keywords for each image/sentence help annotators not deviate too much.
For example, \textit{local keywords} look like \textit{(restaurant, hearing, friends)} $\rightarrow$ \textit{(pictures, menu, difficult)} $\rightarrow$ \textit{(salad, corn, chose)} for three sentences/images, while \textit{global keywords} look like \textit{(wait, salad, restaurant)} for a story of the three sentences/images.

We use Visual Story Telling (ViST)~\cite{visualstorytelling} dataset as our input source.
The dataset contains stories, and each story has five pairs of images and sentences.
We filter out stories that are not temporally ordered using the timestamps of images.
The final number of stories after filtering the non-temporally-ordered stories is 28,130. 
For the denotation experiment, we only use randomly chosen 100 stories.
The detailed pre-processing steps are described in Appendix.



\subsection{Measuring Meaning Preservation \& Style Diversity across Different Denotations}

\begin{figure*}[ht]	
\vspace{0mm}
\centering
{
\includegraphics[trim={1mm 88mm 1mm 0mm},clip,width=.975\linewidth]{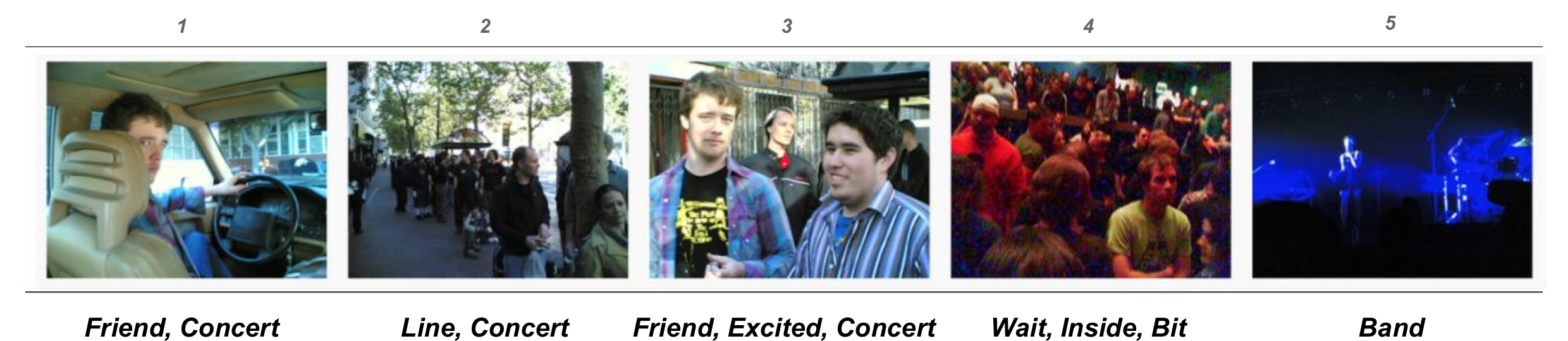}
}
\caption{\label{fig:final_setup} Final denotation setting for data collection: an event that consists of a series of five images with a handful number of keywords. We ask annotators to produce text about the event for each image.\vspace{0mm}}
\end{figure*}

For each denotation setting,
we conduct a quantitative experiment to measure the two metrics: meaning preservation and style diversity.
The two metrics pose a trade-off to each other. The best input setting then is one that can capture both in appropriate amounts. For example, we want meaning of the input preserved, while lexical or syntactic features (e.g., POS tags) can vary depending on annotator's persona.
We use the following automatic measures for the two metrics:

\textbf{Style Diversity} measures how much produced sentences (or stories) differ amongst themselves. 
Higher the diversity, better the stylistic variation in language it contains.
We use an entropy measure to capture the variance of n-gram features between annotated sentences: Entropy (Gaussian-Mixture) that combines the N-Gram entropies \cite{shannon1951prediction} using Gaussian mixture model (N=3).


\textbf{Meaning Preservation} measures semantic similarity of the produced sentence (or story) with the reference sentence (or story).
Higher the similarity, better the meaning preserved.
We use a hard-measure, METEOR \cite{banerjee2005meteor}, that calculates F-score of word overlaps between the output and reference sentences\footnote{Other measures (e.g., BLEU \cite{papineni2002bleu}, ROUGE \cite{lin2003automatic}) show relatively similar performance.}.
Since the hard measures do not take into account all semantic similarities \footnote{METEOR does consider synonymy and paraphrasing but is limited by its predefined model/dictionaries/resources for the respective language, such as Wordnet}, we also use a soft measure, VectorExtrema (VecExt) \cite{liu2016not}. It computes cosine similarity of averaged word embeddings (i.e., GloVe \cite{pennington2014glove}) between the output and reference sentences. 

Table~\ref{tab:denotation} shows results of the two metrics across different input settings we define.
For the sentence level, as expected, \textit{single} \textit{reference} \textit{sentence} has the highest meaning preservation across all the metrics because it is basically paraphrasing the reference sentence.
In general, 
\textit{\textbf{Story}} (\textit{images} + \textit{local} \textit{keywords}) shows a great performance with the highest diversity regardless of the levels, as well as the highest preservation at the soft measure on the story-level.
Thus, we use \textit{\textbf{Story}}(\textit{images}+\textit{local} \textit{keywords}) as the input setting for our final data collection, which has the most balanced performance on both metrics.
Figure~\ref{fig:final_setup} shows an example of our input setting for crowd workers.

\section{\methodpastel: A Parallelly Annotated Dataset for Stylistic Language Dataset}\label{sec:dataset}


We describe how we collect the dataset with human annotations and provide some analysis on it.

\subsection{Annotation Schemes}
Our crowd workers are recruited from the Amazon Mechanical Turk (AMT) platform.
Our annotation scheme consists of two steps: (1) ask annotator's demographic information (e.g., gender, age) and (2) given an input denotation like Figure \ref{fig:final_setup}, ask them to produce text about the denotation with their own style of persona (i.e., connotation).

In the first step, we use seven different types of persona styles;
\textit{gender}, \textit{age}, \textit{ethnic}, \textit{country}, \textit{education} \textit{level}, and \textit{political} \textit{orientation}, and one additional context style \textit{time-of-day} (tod). 
For each type of persona, we provide several categories for annotators to choose.
For example, \textit{political orientation} has three categories: \underline{Centrist}, \underline{Left Wing}, and \underline{Right Wing}.
Categories in other styles are described in the next sub-section.

In the second step, we ask annotators to produce text that describes the given input of denotation.
We again use the pre-processed ViST \cite{visualstorytelling} data in \S\ref{sec:denotation} for our input denotations.
To reflect annotators' persona, we explicitly ask annotators to reflect their own persona in the stylistic writing, instead of pretending others' persona.
We attach detailed annotation schemes at Figure~\ref{fig:annot2} in Appendix.


To amortize both costs and annotators' effort at answering questions, each HIT requires the participants to annotate three stories after answering demographic questions. One annotator was paid \$0.11 per HIT. 
For English proficiency, the annotators were restricted to be from USA or UK. 
A total 501 unique annotators participated in the study. The average number of HIT per annotator was 9.97. 

\begin{table}[ht]
\small
\begin{center}\vspace{0mm}
\caption{\label{tab:dataset} Data statistics of the \methodpastel.\vspace{0mm}}\vspace{0mm}
\begin{tabular}{r|cc}
\toprule
&\textbf{Number of Sentences} & \textbf{Number of Stories} \\\midrule
Train& 33,240&6,648\\
Valid&4,155&831 \\
Test&4,155&831\\\midrule
total&41,550 & 8,310 \\
\bottomrule
\end{tabular}
\end{center}\vspace{0mm}
\end{table}

\begin{table*}[!h]
\centering
\begin{tabularx}{\textwidth}{@{}P{4.6cm}|X@{}}
\toprule
\multicolumn{2}{@{}l}{\textbf{Reference Sentence:} went to an art museum with a group of friends.} \\\midrule 
\textit{edu}:\underline{HighSchoolOrNoDiploma} & My friends and I went to a art museum yesterday . \\\midrule
\textit{edu}:\underline{Bachelor} & I went to the museum with a bunch of friends. \\  
\bottomrule
\end{tabularx}
\newline\vspace*{2mm}\newline
\begin{tabularx}{\textwidth}{@{}P{3.8cm}|X@{}}
\toprule
\multicolumn{2}{@{}l}{\textbf{Reference Sentence:} the living room of our new home is nice and bright with natural light. } \\\midrule
\textit{edu}:\underline{NoDegree}, \textit{gender}:\underline{Male} & The natural lightning made the apartment look quite nice for the upcoming tour .\\\midrule
\textit{edu}:\underline{Graduate}, \textit{gender}:\underline{Female} & The house tour began in the living room which had a sufficient amount of natural lighting. \\ 
\bottomrule 
\end{tabularx}
\newline\vspace*{3mm}\newline
\begin{tabularx}{\textwidth}{@{}P{2.8cm}|X@{}}
\toprule 
\multicolumn{2}{@{}>{\raggedright}p{16cm}}{\textbf{Reference Story:} Went to an art museum with a group of friends . We were looking for some artwork to purchase, as sometimes artist allow the sales of their items . There were pictures of all sorts , but in front of them were sculptures or arrangements of some sort .  Some were far out there or just far fetched . then there were others that were more down to earth and stylish. this set was by far my favorite.very beautiful to me . } \\
\midrule
\textit{edu}:\underline{HighSchool}, \textit{ethnic}:\underline{Caucasian}, \textit{gender}:\underline{Female} & My friends and I went to a art museum yesterday . There were lots of puchases and sales of items going on all day . I loved the way the glass sort of brightened the art so much that I got all sorts of excited . After a few we fetched some grub . My favorite set was all the art that was made out of stylish trash . \\
\midrule
\textit{edu}:\underline{Bachelor}, \textit{ethnic}:\underline{Caucasian}, \textit{gender}:\underline{Female} & I went to the museum with a bunch of friends . There was some cool art for sale . We spent a lot of time looking at the sculptures . This was one of my favorite pieces that I saw . We looked at some very stylish pieces of artwork . \\  
\bottomrule
\end{tabularx}\vspace{0mm}
\caption{Two sentence-level (top, middle) and one story-level (bottom) annotations in \textsc{\methodpastel}.
Each text produced by an annotator has their own persona values (underline) for different types of styles (italic).
Note that the reference sentence (or story) is given for comparison with the annotated text.
Note that misspellings of the text are made by annotators.
\vspace{0mm}
}
\label{tab:sentencestoryExample}
\vspace{0mm}
\end{table*}

Once we complete our annotations, we filter out noisy responses such as stories with missing images and overtly short sentences (i.e., minimum sentence length is 5).
The dataset is then randomly split into train, valid, and test set by 0.8, 0.1, and 0.1 ratios, respectively.
Table~\ref{tab:dataset} shows the final number of stories and sentences in our dataset.

\subsection{Analysis and Examples}

\begin{figure}[ht!]	
\centering\vspace{0mm}
{
\subfloat[][\small{Political orientation}]{\includegraphics[trim={4mm 4mm 4mm 4mm},clip,width=.43\linewidth]{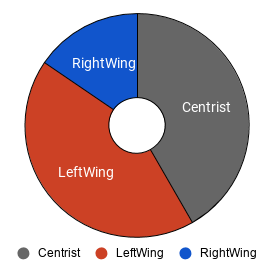}}
\quad
\subfloat[][\small{Gender}]{\includegraphics[trim={2mm 3mm 3mm 4mm},clip,width=.43\linewidth]{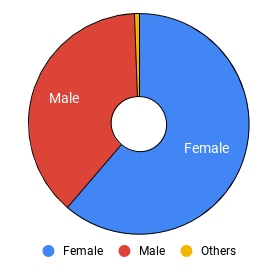}}
\\
\subfloat[][\small{Age}]{\includegraphics[trim={2mm 0mm 4mm 4mm},clip,width=.47\linewidth]{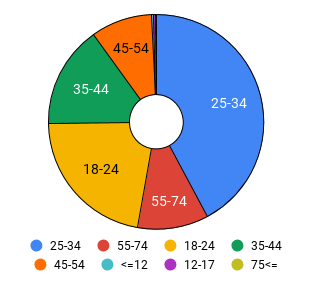}}
\quad
\subfloat[][\small{Education level}]{\includegraphics[trim={4mm 0mm 3mm 3mm},clip,width=.47\linewidth]{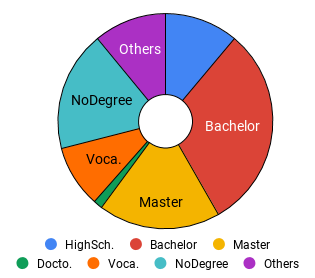}}
}\vspace{0mm}
\caption{\label{fig:dist} Distribution of annotators for each personal style in \methodpastel. Best viewed in color. \vspace{0mm}}
\end{figure}

Figure \ref{fig:dist} shows demographic distributions of the annotators.
Education-level of annotators is well-balanced, while gender and political view are somewhat biased (e.g., 68\% of annotators are Female, only 18.6\% represent themselves as right-wing).
Table~\ref{tab:style_dist} in Appendix includes the categories in other styles and their distributions.

Table~\ref{tab:sentencestoryExample} shows few examples randomly chosen from our dataset: two at sentence level (top, middle) and one at story level (bottom).
Due to paucity of space, we only show a few types of persona styles.
For example, we observe that Education level (e.g., \underline{NoDegree} vs. \underline{Graduate}) actually reflects a certain degree of formality in their writing at both sentence and story levels.
In \S\ref{sec:classification}, we conduct an in-depth analysis of textual variation with respect to the persona styles in \methodpastel.

\section{Applications with \methodpastel}\label{sec:exp}

\methodpastel can be used in many style related  applications including style classification, stylometry~\cite{verhoeven2014clips}, style transfer~\cite{fu2017style}, visually-grounded style transfer, and more.
Particularly, we chose two applications, where \methodpastel helps design appropriate experiment and evaluation: controlled style classification (\S\ref{sec:classification}) and supervised style transfer (\S\ref{sec:transer}).

\subsection{Controlled Style Classification}\label{sec:classification}

A common mistake in style classification datasets is not controlling external style variables when predicting the category of the target style.
For example, when predicting a gender type given a text $P$($\textit{gender}$=$\underline{Male}$$\mid$$\texttt{text})$, the training data is only labeled by the target style \textit{gender}.
However, the \texttt{text} is actually produced by a person with not only $\textit{gender}$=$\underline{Male}$ but also other persona styles such as \textit{age}=\underline{55-74} or \textit{education}=\underline{HighSchool}.
Without controlling the other external styles, the classifier is easily biased against the training data.

We define a task called \textit{controlled style classification} where all other style variables are fixed\footnote{The distribution of number of training instances per variable is given in Appendix}, except one to classify. 
Here we evaluate (1) which style variables are relatively difficult or easy to predict from the text given, and (2) what types of textual features are salient for each type of style classification.


\paragraph{Features.}
Stylistic language has a variety of features at different levels such as lexical choices, syntactic structure and more. 
Thus, we use following features:
\begin{itemize}[noitemsep,topsep=0pt,leftmargin=*]
\item \textbf{lexical} features: ngram's frequency (n=3), number of named entities, number of stop-words
\item \textbf{syntax} features: sentence length, number of each Part-of-Speech (POS) tag, number of out-of-vocabulary, number of named entities 
\item \textbf{deep} features: pre-trained sentence encoder using BERT \cite{devlin2018bert} 
\item \textbf{semantic} feature: sentiment score 
\end{itemize}
where named entities, POS tags, and sentiment scores are obtained using the off-the-shelf tools such as Spacy\footnote{\url{https://spacy.io/}} library. 
We use $70K$ n-gram lexical features, $300$ dimensional embeddings, and $14$ hand-written features.

\begin{figure}[ht]	
\centering\vspace{0mm}
{
\subfloat[][\small{different style types}]{\includegraphics[trim={0 0 0 0},clip,width=.98\linewidth]{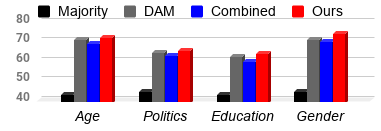}}\vspace{0mm}
\\
\subfloat[][\small{sentences vs stories}]{\includegraphics[trim={0 0 0 0},clip,width=.98\linewidth]{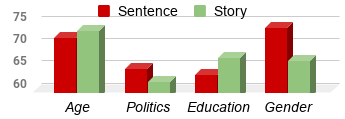}}
}
\caption{\label{fig:controlled_exp} Controlled style classification: F-scores on (a) different types of styles on sentences and on (b) our best models between sentences and stories. Best viewed in color.\vspace{0mm}}\vspace{0mm}
\end{figure}

\paragraph{Models.}
We train a binary classifier for each personal style with different models: logistic regression, SVM with linear/RBF kernels, Random Forest, Nearest Neighbors, Multi-layer Perceptron, AdaBoost, and Naive Bayes.
For each style, we choose the best classifiers on the validation.
Their F-scores are reported in Figure \ref{fig:controlled_exp}. 
We use sklearn's implementation of all models \cite{pedregosa2011scikit}.\footnote{\url{http://scikit-learn.org/stable/}}
We consider various regularization parameters for SVM and logistic regression (e.g., c=[0.01, 0.1, 0.25, 0.5, 0.75, 1.0].

We use neural network based baseline models: deep averaging networks \cite[DAN,][]{Iyyer:2015} of GloVe word embeddings \cite{pennington2014glove}\footnote{Other architectures such as convolutional neural networks \cite[CNN,][]{Zhang:2015} and recurrent neural networks \cite[LSTM,][]{Hochreiter:1997} show comparable performance as DAN.}.
We also compare with the non-controlled model (Combined) which uses a combined set of samples across all other variables except for one to classify using the same features we used.

\paragraph{Setup.}
We tune hyperparameters using 5-fold cross validation.
If a style has more than two categories, we choose the most conflicting two: \textit{gender}:\{\underline{Male}, \underline{Female}\}, 
\textit{age}: \{\underline{18-24}, \underline{35-44}\},
\textit{education}: \{\underline{Bachelor}, \underline{No} \underline{Degree}\},
and \textit{politics}: \{\underline{LeftWing}, \underline{RightWing}\}. 
To classify one style, all possible combinations of other styles ($2*2*2$=$8$) are separately trained by different models. 
We use the macro-averaged F-scores among the separately trained models on the same test set for every models.



\paragraph{Results.}

Figure~\ref{fig:controlled_exp} shows F-scores (a) among different styles and (b) between sentences and stories.
In most cases, multilayer perceptron (MLP) outperforms the majority classifier and other models by large margins.
Compared to the neural baselines and the combined classifier, our models show better performance.
In comparison between controlled and combined settings, controlled setting achieves higher improvements, indicating that fixing external variables helps control irrelevant features that come from other variables.
Among different styles, gender is easier to predict from the text than ages or education levels.
Interestingly, a longer context (i.e., story) is helpful in predicting age or education, whereas not for political view and gender.

In our ablation test among the feature types, the combination of different features (e.g., lexical, syntax, deep, semantic) is very complementary and effective.
Lexical and deep features are two most significant features across all style classifiers, while syntactic features are not.


\begin{table}[ht]
\small\vspace{0mm}
\begin{center}
\caption{\label{tab:style_salient} Most salient lexical (lower cased) and syntactic (upper cased) features on story-level classification. Each feature is chosen by the highest coefficients in the logistic regression classifier.\vspace{0mm}
}
\centering
\begin{tabularx}{\linewidth}{@{}P{3.7cm}|P{3.7cm}@{}}
\toprule
\textit{Gender}:\underline{Male} & \textit{Gender}:\underline{Female} \\\midrule
PROPN, ADJ, \#\_ENTITY, went, party, SENT\_LEN
&
happy, day, end, group, just, snow, NOUN
\\\midrule
\toprule
\textit{Politics}:\underline{LeftWing} & \textit{Politics}:\underline{RightWing} \\\midrule
female, time, NOUN, ADP, VERB, porch, day, loved
&
SENT\_LENGTH, PROPN, \#\_ENTITY, n't, ADJ, NUM
\\\midrule
\toprule
\textit{Education}:\underline{Bachelor} & \textit{Education}:\underline{NoDegree} \\\midrule
food, went, \#\_STOPWORDS, race, ADP
&
!, just, came, love, lots, male, fun, n't, friends, happy
\\\midrule
\toprule
\textit{Age}:\underline{18-24} & \textit{Age}:\underline{35-44} \\\midrule
ADP, come, PROPN, day, ride, playing, sunset
&
ADV, did, town, went, NOUN, \#\_STOPWORDS
\\
\bottomrule
\end{tabularx}
\end{center}
\vspace{0mm}
\end{table}

Table~\ref{tab:style_salient} shows the most salient features for classification of each style.
Since we can't interpret deep features, we only show lexical and syntactic features.
The salience of features are ranked by coefficients of a logistic regression classifier.
Interestingly, female annotators likely write more nouns and lexicons like `happy', while male annotators likely use pronouns, adjectives, and named entities.
Annotators on left wing prefer to use `female', nouns and adposition, while annotators on right wing prefer shorter sentences and negative verbs like `n't'.
Not many syntactic features are observed from annotators without degrees compared to with bachelor degree. 

\subsection{Supervised Style Transfer}\label{sec:transer}

The style transfer is defined as \textsc{($S$, ${\alpha}$) $\rightarrow$ ${\hat{S}}$}:
We attempt to alter a given source sentence $S$ to a given target style ${\alpha}$. The model generates a candidate target sentence ${\hat{S}}$ which preserves the meaning of $S$ but is more faithful to the target style ${\alpha}$ so being similar to the target annotated sentence ${\thickbar{S}_\alpha}$. 
We evaluate the model by comparing the predicted sentence ${\hat{S}}$ and target annotated sentence ${\thickbar{S}_\alpha}$.
The sources are from the original reference sentences, while the targets are from our annotations.

\paragraph{Models.}
We compare five different models:
\begin{itemize}[noitemsep,topsep=0pt,leftmargin=*]
    \item \textsc{AsItIs}: copies over the source 
    sentence to the target, without any alterations.
    \item \textsc{WordDistRetrieve}: retrieves a training source-target pair that has the same target style as the test pair and is closest to the test source in terms of word edit distance \cite{navarro2001guided}. It then returns the target of that pair.
     \item \textsc{EmbDistRetrieve}: Similar to \textsc{WordDistRetrieve}, except that a continuous bag-of-words (CBOW) is used to retrieve closest source sentence instead of edit distance.
     \item \textsc{Unsupervised}: use unsupervised style transfer models using Variational Autoencoder \cite{shen2017style} and using additional objectives such as cross-domain and adversarial losses \cite{lample2017unsupervised}\footnote{We can't directly compare with \citet{hu2017toward,prabhumoye2018style} since their performance highly depends on the pre-trained classifier that often shows poor performance.}.
     Since unsupervised models can't train multiple styles at the same time, we train separate models for each style and macro-average their scores at the end.
     In order not to use the parallel text in \methodpastel, we shuffle the training text of each style.
     \item \textsc{Supervised}: uses a simple attentional sequence-to-sequence (S2S) model \cite{bahdanau2014neural} extracting the parallel text from \methodpastel. The model jointly trains different styles in conjunction by concatenating them to the source sentence at the beginning.
\end{itemize}
We avoid more complex architectural choices for \textsc{Supervised} models like adding a pointer component or an adversarial loss, since we seek to establish a minimum level of performance on this dataset.

\paragraph{Setup.}
We experiment with both \textsc{Softmax} and \textsc{Sigmoid} non-linearities to normalize attention scores in the sequence-to-sequence attention. 
Adam \cite{kingma2014adam} is used as the optimizer. Word-level cross entropy of the target is used as the loss. The batch size is set to $32$. We pick the model with lowest validation loss after 15 training epochs. All  models are implemented in PyTorch \cite{paszke2017pytorch}.

For an evaluation, in addition to the same hard and soft metrics used for measuring the meaning preservation in \S\ref{sec:denotation}, we also use BLEU$_2$ \cite{papineni2002bleu} for unigrams and bigrams, and ROUGE \cite{lin2003automatic} for hard metric and Embedding Averaging (EA) similarity \cite{liu2016not} for soft metric.

\begin{table}[ht]
\begin{center}
\small\vspace{0mm}
\caption{\label{tab:StyleTransferStats} Supervised style transfer. 
\textsc{GloVe} initializes with pre-trained word embeddings.
\textsc{PreTr.} denotes pre-training on YAFC. 
Hard measures are \textbf{B}LEU$_{\bf 2}$, \textbf{M}ETEOR, and \textbf{R}OUGE, and soft measures are \textbf{E}mbeding\textbf{A}veraging and \textbf{V}ector\textbf{E}xtrema.\vspace{0mm}
}\small\vspace{0mm}
\centering
\begin{tabular}{@{}l@{}|@{}P{0.9cm}@{}P{0.9cm}@{}P{0.9cm}@{}|@{}P{1cm}@{}P{1cm}@{}}
\toprule
 &  \multicolumn{3}{@{}c|}{Hard (${\hat{S}}$,${\thickbar{S}_\alpha}$)} &  \multicolumn{2}{c@{}}{Soft (${\hat{S}}$,${\thickbar{S}_\alpha}$)}   \\\midrule
Models: \textsc{($S$, ${\alpha}$) $\rightarrow$ ${\hat{S}}$} & \textbf{B}$_2$ & \textbf{M} & \textbf{R} & \textbf{EA} & \textbf{VE} \\
\midrule
\textsc{AsItIs} & \textbf{35.41} & \textbf{12.38} & 21.08 & 0.649 & 0.393 \\ 
\textsc{WordDistRetrieve} & 30.64  & 7.27 & 22.52 & 0.771 & 0.433 \\
\textsc{EmbDistRetrieve} & 33.00 & 8.29 & 24.11 & 0.792 & 0.461 \\ 
\hline
\multicolumn{5}{@{}l}{\textsc{Unsupervised}}\\
 $\cdot$ \citet{shen2017style} & 23.78 & 7.23 & 21.22 & 0.795 & 0.353\\
 $\cdot$ \citet{lample2017unsupervised} & 24.52	&6.27&	19.79	&	0.702&	0.369\\
\midrule
\multicolumn{5}{@{}l}{\textsc{Supervised}}\\
 $\cdot$ \textsc{S2S} & 26.78  & 7.36 & 25.57 & 0.773 & 0.455  \\
 $\cdot$ \textsc{S2S+GloVe} & 31.80  & 10.18 & \underline{29.18} & \underline{0.797} & \underline{0.524}  \\
 $\cdot$ \textsc{S2S+GloVe+PreTr.} & \underline{31.21}  & \underline{10.29} & \textbf{29.52} & \textbf{0.804} & \textbf{0.529}  \\
\bottomrule
\end{tabular}
\end{center}
\vspace{0mm}
\end{table}

\begin{table*}[!h]
\centering
\addtolength{\tabcolsep}{-1pt}
\vspace{0mm}
\begin{tabularx}{\textwidth}{@{}X@{}}
\toprule
\textbf{Source ($S$)}: I'd never seen so many beautiful flowers.\\
\midrule
\textbf{Style (\textsc{$\alpha$})}: (\underline{Morning}, \underline{HighSchool})\\
\quad $S$ + $\alpha$ $\rightarrow$ \textsc{$\hat{S}$}: the beautiful flowers were beautiful. \\
\quad $\thickbar{S}_\alpha$: the flowers were in full bloom.  \\
\textbf{Style (\textsc{$\alpha$})}: (\underline{Afternoon}, \underline{NoDegree})\\
\quad $S$ + $\alpha$ $\rightarrow$ \textsc{$\hat{S}$}: The flowers were very beautiful. \\
\quad $\thickbar{S}_\alpha$: Tulips are one of the magnificent varieties of flowers.\\
\bottomrule
\end{tabularx}
\newline \vspace*{3mm}\newline
\begin{tabularx}{\textwidth}{@{}X@{}}
\toprule
\textbf{Source ($S$)}: she changed dresses for the reception and shared food with her new husband.\\
\midrule
\textbf{Style (\textsc{$\alpha$})}: (\underline{Master}, \underline{Centrist})\\
\quad $S$ + $\alpha$ $\rightarrow$ \textsc{$\hat{S}$}: The woman had a great time with her husband\\
\quad $\thickbar{S}_\alpha$: Her husband shared a cake with her during reception\\
\midrule
\textbf{Style (\textsc{$\alpha$})}: (\underline{Vocational}, \underline{Right})\\
\quad $S$ + $\alpha$ $\rightarrow$ \textsc{$\hat{S}$}: The food is ready for the reception \\
\quad $ \thickbar{S}_\alpha$: The new husband shared the cake at the reception\\
\bottomrule
\end{tabularx}\small\vspace{0mm}
\caption{Examples of style transferred text by our supervised model (\textsc{S2S+GloVe+PreTr.}) on \textsc{\methodpastel}. Given source text ($S$) and style ($\alpha$), the model predicts a target sentence $\hat{S}$ compared to annotated target sentence $\thickbar{S}_\alpha$.\vspace{0mm}
}
\label{tab:QualitativeDatasetTestExamples}
\vspace{0mm}\end{table*}


\paragraph{Results.}
Table \ref{tab:StyleTransferStats} shows our results on style tranfer.
We observe that initializing both en/decoder's word embeddings with \textsc{GloVe} \cite{pennington2014glove} improves model performance on most metrics. 
Pretraining (\textsc{PreTr.}) on the formality style transfer data \textsc{YAFC} \cite{rao2018dear} further helps performance.
All supervised \textsc{S2S} approaches outperform both retrieval-based baselines on all measures. This illustrates that the performance scores achieved are not simply a result of memorizing the training set. 
\textsc{S2S} methods surpass \textsc{AsItIs} on both soft measures and ROUGE. The significant gap that remains on BLEU remains a point of exploration for future work.
The significant improvement against the unsupervised methods \cite{shen2017style,lample2017unsupervised} indicates the usefulness of the parallel text in \methodpastel.

Table~\ref{tab:QualitativeDatasetTestExamples} shows output text ${\hat{S}}$ produced by our model given a source text $S$ and a style ${\alpha}$.
We observe that the output text changes according to the set of styles.

\section{Conclusion and Future Directions}
We present \textsc{\methodpastel}, a parallelly annotated stylistic language dataset. 
Our dataset is collected by human annotation using a proper denotation setting that preserves the meaning as well as maximizes the diversity of styles.
Multiplicity of persona styles in \methodpastel makes other style variables controlled (or fixed) except the target style for classification, which is a more accurate experimental design.
Our simple supervised models with our parallel text in \methodpastel outperforms the unsupervised style transfer models using non-parallel corpus. 
We hope \methodpastel can be a useful benchmark to both train and evaluate models for style transfer and other related problems in text generation field. 


We summarize some directions for future style researches:
\begin{itemize}[noitemsep,topsep=0pt,leftmargin=*]
\item In our ablation study, salient features for style classification are not only syntactic or lexical features but also content words (e.g., love, food). This is a counterexample to the hypothesis implicit in much of recent style research: \textit{style} needs to be \textit{separately modeled} from \textit{content}. We also observe that some texts remain similar across different annotator personas or across outputs from our transfer models, indicating that some content is stylistically invariant. Studying these and other aspects of the  content-style relationship in \methodpastel could be an interesting direction.
\item Does any external variable co-varying with the text qualify to be a style variable/facet? What are the categories of style variables/facets? Do architectures which transfer well across one style variable (e.g gender) generalize to other style variables (e.g age)? We opine that these questions are largely overlooked by current style transfer work. We hope that our consideration of some of these questions in our work, though admittedly rudimentary, will lead to them being addressed  extensively in future style transfer work.
\end{itemize}

\section*{Acknowledgements}
This work would not have been possible without the ViST dataset and helpful suggestions with Ting-Hao Huang.
We also thank Alan W Black, Dan Jurafsky, Wei Xu, Taehee Jung, and anonymous reviewers for their helpful comments.

\bibliography{pastel}
\bibliographystyle{acl_natbib}

\clearpage
\renewcommand*\appendixpagename{\Large Appendices}
\begin{appendix}

\section{Data Collection: Details}


Here we describe additional details that we face during our data collection.
There are some technical difficulties we faced during annotations:
\begin{itemize}[noitemsep,topsep=0pt,leftmargin=*]
\item 
Since it is nontrivial to verify whether the annotator  had understood our instructions insufficiently, we filter out very short sentence annotations.
\item Some images from \textit{VIST} were too heavy to load or unavailable. Most annotators mark "No Image", "Image not Loaded" etc for such images. We filter out such annotations using hand-coded filtering rules.
\item Some annotators exploit the large number of HITs available and difficulty of verification to game the study and perform  large number of HITs carelessly. We manually block such annotators after a minimum threshold.
\item We increase pay from $0.07$ to $0.11$ based on annotator feedback.
\end{itemize}
Finally, we filter out 4.6\% of noisy annotations.

\begin{figure*}[h!]	
\centering
\small
{
\subfloat[][\normalsize{Annotation scheme for asking persona w.r.t. given denotation with their persona.}]{\includegraphics[trim={0 0 0 0},clip,width=.9\linewidth]{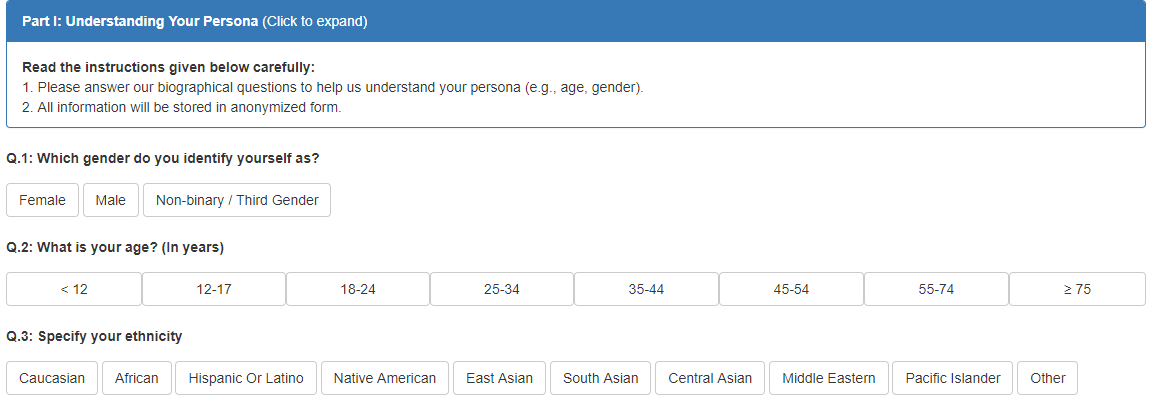}}
\\\vspace{3mm}
\subfloat[][\normalsize{Annotation scheme for generating sentences}]{\includegraphics[trim={0 0 0 0},clip,width=.9\linewidth]{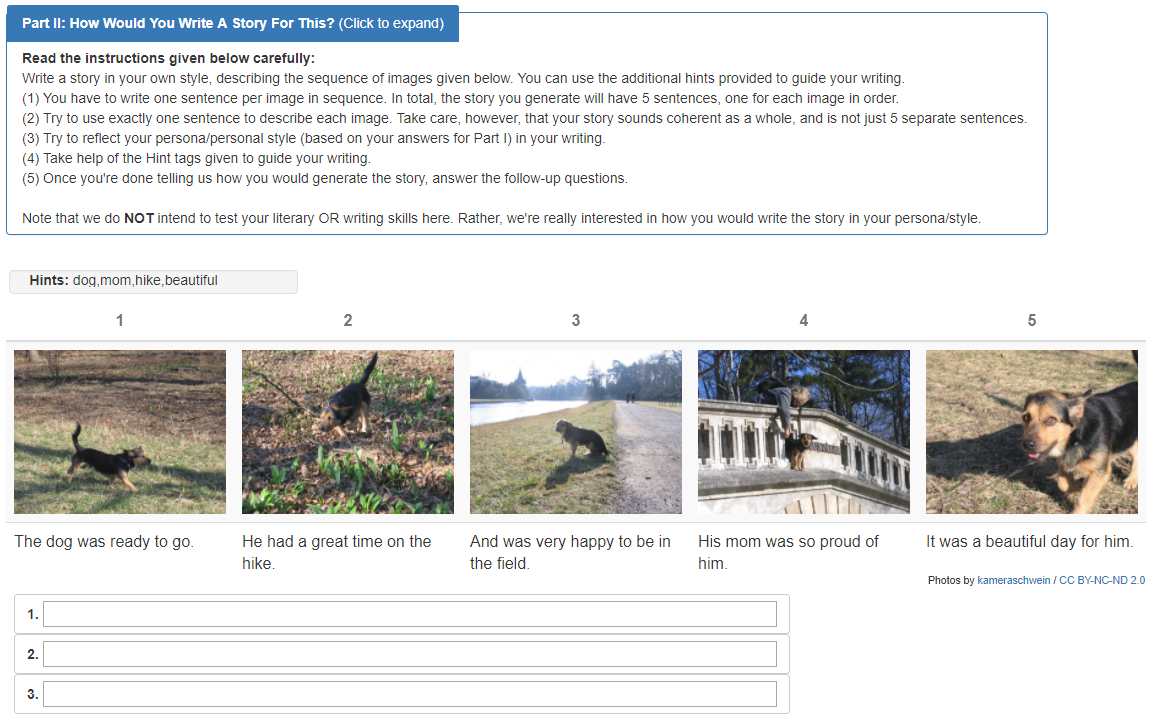}}
}
\caption{\label{fig:annot2} A part of our annotation schemes for asking annotators to generate sentences given a sequence of stories, keywords, and reference sentences. }
\end{figure*}
Figure~\ref{fig:annot2} shows a part of our annotation schemes for asking annotators to generate sentences given a sequence of stories, keywords, and reference sentences.

\section{\methodpastel: Details}

\begin{table}[h]
\begin{center}
\small
\caption{\label{tab:style_dist} Distribution of stories.}\vspace{0mm}
\centering
\begin{tabularx}{\linewidth}{@{}p{1.1cm}@{}X@{}}
\toprule
\textbf{Style} & \textbf{Categories with number of stories annotated}\\
\midrule
gender &   Male (1320), Female (2908), Others (45) \\
\midrule
age & $<$12 (4), 12-17 (1), 18-24 (1050), 25-34 (1527), 35-44 (720), 45-54 (475), 55-74 (493), 75$<$ (3)  \\
\midrule
ethnic  & Hispanic/Latino (81), MiddleEastern (4), SouthAsian (27), NativeAmerican (115), PacificIslander (26), CentralAsian (3), African (270), EastAsian (112), Caucasian (3521), Other (114)\\
\midrule
country  & UK (98), USA (4172), Others (3) \\
\midrule
edu & Associate (423), TechTraining (488), NoSchool (3), HighSchoolOrNoDiploma (47), Bachelor (1321), Master (440), NoDegree (939), HighSchool (557), Doctorate (55) \\
\midrule
politics  & Centrist (1786), LeftWing (1691), RightWing (796) \\
\midrule
time & Morning (1103), Evening (392), Midnight (282), Afternoon (1666), Night (830)\\
\bottomrule
\end{tabularx}
\end{center}\vspace{0mm}
\end{table}


Table~\ref{tab:style_dist} shows distribution of stories for each category of styles.

\begin{table}[ht!]
\begin{center}\vspace{0mm}
\caption{\label{tab:qualityStats} Quality statistics of \methodpastel. 
\vspace{0mm} }
\centering
\small
\begin{tabular}{@{}l|P{0.5cm}P{0.6cm}|P{0.6cm}P{0.6cm}@{}}
\toprule
& \textbf{Sent} & \textbf{Story} & \textbf{Quora}     \\ 
\midrule
Avg. Preservation & 2.69 & 3.89 & 3.72 \\ 
Avg. Fluency & 4.02 & 3.46 & 4.07 \\ 
Highly Preserved \% & 45.2 & 77.4 & 83.5 \\ 
Highly Preserved\&Fluent \% & 44.6    & 76.3 & 80.1 \\ 
\bottomrule
\end{tabular}
\end{center}
\vspace{0mm}
\end{table}

We also conduct add \textit{Meaning Preservation} human study to evaluate the quality of annotations. 
We recruit different annotators from the data collection and ask them to rate the two following questions under 5 point scale: (1) Does an annotated sentence $S_{2}$ from our dataset preserve the meaning of the original reference sentence $S_{1}$? and (2) How fluent $S_{2}$ is? We sample three annotations each for $1000$ sentences and $800$ stories from our test dataset. For a scale-calibration, we also annotate Quora \cite{iyer2017first} by asking annotators to evaluate the Quora pairs as a control. 
Table \ref{tab:qualityStats} summarizes the results. 
In detail, we see that the story-level annotations keep more meaning presentation, while the sentence-level ones hold more fluency. 


\section{Controlled Style Classification: Details}


\begin{table*}[h!]
\vspace{0mm}
\small
\begin{center}
\caption{\label{tab:label_dist} Label distribution for each combination of external variables in Controlled Style classification.
}
\centering
\begin{tabular}{@{}l@{}|@{}c@{}c@{}}
\toprule
target: \textbf{politics} & LeftWing & RightWing \\\hline
\quad (Male, 18-24, NoDegree) & 59 & 206 \\
\quad (Female, 18-24, Bachelor) & 147 & 42 \\
\quad (Female, 18-24, NoDegree) & 95 & 15 \\
\quad (Male, 35-44, Bachelor) & 21 & 4 \\
\quad (Male, 18-24, Bachelor) & 18 & 11 \\
\quad (Female, 35-44, Bachelor) & 63 & 13 \\
\quad (Female, 35-44, NoDegree) & 24 & 7 \\
\quad (Male, 35-44, NoDegree) & - & 12 \\
\hline
target: \textbf{education} & NoDegree & Bachelor \\\hline
\quad (Male, 35-44, RightWing) & 12 & 4 \\
\quad (Female, 18-24, LeftWing) & 95 & 147 \\
\quad (Female, 35-44, RightWing) & 7 & 13 \\
\quad (Male, 35-44, LeftWing) & - & 21  \\
\quad (Male, 18-24, LeftWing) & 59 & 18 \\
\quad (Female, 18-24, RightWing) & 15 & 42 \\
\quad (Female, 35-44, LeftWing) & 24 & 63 \\
\quad (Male, 18-24, RightWing) & 206 & 11 \\
\bottomrule
\end{tabular}
\quad\quad
\begin{tabular}{@{}l@{}|@{}c@{}c@{}}
\toprule
target: \textbf{age} & 18-24 & 35-44 \\\hline
\quad (Female, NoDegree, LeftWing) & 95 & 24 \\
\quad (Male, Bachelor, RightWing) & 11 & 4 \\
\quad (Female, Bachelor, LeftWing) & 147 & 63 \\
\quad (Male, NoDegree, LeftWing) & 59 & -  \\
\quad (Female, NoDegree, RightWing) & 15 & 7 \\
\quad (Male, NoDegree, RightWing) & 206 & 12 \\
\quad (Male, Bachelor, LeftWing) & 18 & 21 \\
\quad (Female, Bachelor, RightWing) & 42 & 13 \\
\hline
target: \textbf{gender} & Male  & Female \\\hline
\quad (18-24, Bachelor, LeftWing) & 18 & 147 \\
\quad (35-44, NoDegree, RightWing) & 12 & 7 \\
\quad (18-24, NoDegree, LeftWing) & 59 & 95 \\
\quad (35-44, Bachelor, LeftWing) & 21 & 63  \\
\quad (35-44, NoDegree, LeftWing) &  -& 24 \\
\quad (35-44, Bachelor, RightWing) & 4 & 13 \\
\quad (18-24, Bachelor, RightWing) & 11 & 42 \\
\quad (18-24, NoDegree, RightWing) & 206 & 15\\
\bottomrule
\end{tabular}
\end{center}\vspace{0mm}
\end{table*}
For controlled setting, Table~\ref{tab:label_dist} shows label distribution for each combination of external variables.
We calculate the final f-scores of each style by macro averaging the combinations.
If a combination has zero number of instances for one type of class value, we ignore them in the calculation.

Table~\ref{tab:more_example_sentence} and Table~\ref{tab:more_example_story} show randomly chosen examples from the validation split of our dataset.
It shows reference sentence and some sentences written by our annotators, along with their associated demographic traits.
Table \ref{tab:QualitativeDatasetTestExamplesComplete} has more model outputs in style tranfer.

\begin{table*}[!h]
\centering
\small
\begin{tabularx}{\textwidth}{@{}c|P{4cm}|X@{}}
\toprule
No & Style  & Text \\ 
\midrule
\multirow{3}{3mm}{1} &   \textit{Reference} & went to an art museum with a group of friends . \\
&  \underline{HighSchool} & My friends and I went to a art museum yesterday . \\
& \underline{Bachelor} & I went to the museum with a bunch of friends . \\  
\midrule
\multirow{3}{3mm}{2} &  \textit{Reference} & dad always said if you misbehave, you will end up in the local jail \\
&  \underline{British, HighSchool, Female} & misbehaving people sent to local jail .  \\
& \underline{American, Graduate, Female} & They misbehave at the local jail . \\  
\midrule
\multirow{3}{3mm}{3} 
&  \textit{Reference} & This one was really cool as it rolled down the hill with the people in it.  \\
&  \underline{American, Graduate, Female} & Rolling together is a messy, but bonding experience. \\
& \underline{American, Graduate, Male} & He rolled with other people down the hill. \\ 
\midrule
\multirow{3}{3mm}{4}
&  \textit{Reference} & in the woods, there was a broken tree  .  \\
 &  \underline{Female} & A broken tree stands in the center of a path in the woods . \\
& \underline{Male} & Doing my daily walk in the fores and I decided to go a different path and look what I came across , a tree in the middle of a path , interesting . \\  
\midrule
\multirow{3}{3mm}{5} 
& \textit{Reference} & the vehicles were lined up to marvel at .  \\
&  \underline{Caucasian, Doctoral} & I went to the car show. \\
& \underline{NativeAmerican, Masters} & The car is the super.  \\  
\midrule
\multirow{3}{3mm}{6} 
&  \textit{Reference} & we went to a concert last night .  \\
&  \underline{VocationalTraining} & The concert we went to last night was great . \\
& \underline{AssociateDegree} & I have been looking forward to this concert all year . \\  

\bottomrule
\end{tabularx}
\caption{Sentence examples from the validation split of our dataset showing the reference sentence and some sentences written by our annotators, along with their associated demographic traits.}
\label{tab:more_example_sentence}
\end{table*}

\begin{table*}[h!]
\centering
\small
\begin{tabularx}{\textwidth}{@{}c@{}|P{3.5cm}|X@{}}
\toprule 
No & Style  & Text \\ 
\midrule
\multirow{10}{3mm}{1}
&  \textit{Reference} & Went to an art museum with a group of friends . We were looking for some artwork to purchase, as sometimes artist allow the sales of their items . There were pictures of all sorts , but in front of them were sculptures or arrangements of some sort .  Some were far out there or just far fetched . then there were others that were more down to earth and stylish. this set was by far my favorite.very beautiful to me .  \\
 &  \underline{Caucasian}, \underline{Female}, \underline{HighSchool} & My friends and I went to a art museum yesterday . There were lots of puchases and sales of items going on all day . I loved the way the glass sort of brightened the art so much that I got all sorts of excited . After a few we fetched some grub . My favorite set was all the art that was made out of stylish trash . \\

& \underline{Caucasian}, \underline{Female}, \underline{Bachelor} & I went to the museum with a bunch of friends . There was some cool art for sale . We spent a lot of time looking at the sculptures . This was one of my favorite pieces that I saw . We looked at some very stylish pieces of artwork . \\  
\midrule
\multirow{10}{3mm}{2}
&  \textit{Reference} & We went on a hiking tour as a family .
  We took a break so that we could hydrate and get something to eat .
  We were so glad to see the finish line of the hike .
  We were heading to the lodging for our overnight stay .
  The scenary was so beautiful on the tour . \\

 &  \underline{Caucasian}, \underline{Female}, \underline{Graduate}, \underline{Night} & This weekend , we took a hiking tour as a family . We had to remember to take a break to eat and hydrate . We hiked in a line , and I 'm glad we did . We decided to stay overnight in the provided lodging . The scenery of the tour was beautiful . \\

& \underline{Caucasian}, \underline{Female}, \underline{Graduate}, \underline{Evening} & Our family took a hiking tour along the blue ridge parkway . We took plenty of water and snacks so we could hydrate and eat on breaks . We were glad to hike in a line up the hill . We chose to stay overnight in some local lodging . We loved the tour and its beautiful scenery . \\  
\bottomrule
\end{tabularx}
\caption{Story examples from the validation split of our dataset showing the reference sentence and some sentences written by our annotators, along with their associated demographic traits.}
\label{tab:more_example_story}
\end{table*}

\begin{table*}[htbp]
\centering
\small
\begin{tabularx}{\textwidth}{@{}l|P{3.8cm}|X@{}}
\toprule
\textbf{No} & \textbf{Type}  & \textbf{Text} \\
\midrule
\multirow{3}{*}{1} &   \textit{Reference} & the guy had buckets of food \\
&  \underline{Bachelor}, \underline{RightWing} & Before they headed out for Spring Break, Joe went out and brought back food for the gang. \\ 
& \underline{HighSchool}, \underline{LeftWing} &  In college , Mary knew a guy named Greg who always ate his food from buckets .   \\
& \textsc{M}: \underline{Bachelor}, \underline{RightWing} & The food was filled with food. \\
& \textsc{M}: \underline{HighSchool}, \underline{LeftWing} & The food was filled with a lot of food . \\ 
\midrule
\multirow{3}{*}{2} &  \textit{Reference} & the bride and groom walked down the aisle together . \\
&  \underline{Midnight}, \underline{Centrist} & The bride and groom walked out after the ceremony .  \\
& \underline{Night}, \underline{RightWing} & The bride and groom happily walked down the aisle after being pronounced ``Man and Wife '' .   \\  
& \textsc{M}: \underline{Midnight}, \underline{Centrist} & The bride and groom were ready to be married . \\
& \textsc{M}: \underline{Night}, \underline{RightWing} & The bride and groom were ready to be married .   \\ 
\midrule
\multirow{3}{*}{3} 
&  \textit{Reference} & we picked up our breakfast from a street vendor selling fresh fruits . \\
&  \underline{NoDegree} & Along the way , you could see vendors on the street . \\
& \underline{Bachelor} & We loved the market stands .  \\  
& \textsc{M}: \underline{NoDegree} & The family went to a restaurant . \\
& \textsc{M}: \underline{Bachelor} & The family went to a restaurant for some delicious food . \\ 
\midrule
\multirow{3}{*}{4}
&  \textit{Reference} & I'd never seen so many beautiful flowers.  \\
 &  \underline{Morning}, \underline{HighSchool} & the flowers were in full bloom. \\
& \underline{Afternoon}, \underline{NoDegree} & Tulips are one of the magnificent varieties of flowers.    \\ 
& \textsc{M}: \underline{Morning}, \underline{HighSchool} & the beautiful flowers were beautiful. \\
& \textsc{M}: \underline{Afternoon}, \underline{NoDegree} & The flowers were very beautiful. \\
\bottomrule
\end{tabularx}
\caption{Examples from \textit{\methodpastel}, along with model outputs of the best performing style transfer model. We show the Reference sentence (\textit{Reference}) and some sentences written by  annotators. We only show the style attributes that differ between the annotators. \textit{\textsc{M}:} denotes the model output given the corresponding Reference as source sentence along with the same target style.
}
\label{tab:QualitativeDatasetTestExamplesComplete}
\end{table*}

\end{appendix}

\end{document}